\documentclass{article}

\usepackage[square,numbers]{natbib}
\usepackage[preprint]{neurips_2020}

\usepackage{amsmath}
\usepackage[utf8]{inputenc} 
\usepackage[T1]{fontenc}    
\usepackage{hyperref}       
\usepackage{url}            
\usepackage{booktabs}       
\usepackage{amsfonts}       
\usepackage{nicefrac}       
\usepackage{microtype}      
\usepackage[ruled,linesnumbered]{algorithm2e}
\usepackage[pdftex]{graphicx} 
\usepackage{caption}
\usepackage{subcaption}

\newcommand{\var}{\texttt}

\title{Learning Functions to Study the Benefit of Multitask Learning\\
}

\author{%
  Gabriele Bettgenhäuser\\
  Spoken Language Systems (LSV)\\
  Saarland University\\
  Saarbrücken, 66123 Germany \\
  \texttt{gbettgenhauser@lsv.uni-saarland.de} \\
  \And
  Michael A. Hedderich\\
  Spoken Language Systems (LSV)\\
  Saarland University\\
  Saarbrücken, 66123 Germany \\
  \texttt{mhedderich@lsv.uni-saarland.de} \\
  \And
  Dietrich Klakow\\
  Spoken Language Systems (LSV)\\
  Saarland University\\
  Saarbrücken, 66123 Germany \\
  \texttt{Dietrich.Klakow@lsv.uni-saarland.de} \\
}

\begin{document}
\bibliographystyle{unsrtnat}
\maketitle

\begin{abstract}
We study and quantify the generalization patterns of multitask learning (MTL) models for sequence labeling tasks. MTL models are trained to optimize a set of related tasks jointly. Although multitask learning has achieved improved performance in some problems, there are also tasks that lose performance when trained together. These mixed results motivate us to study the factors that impact the performance of MTL models. We note that theoretical bounds and convergence rates for MTL models exist, but they rely on strong assumptions such as task relatedness and the use of balanced datasets. To remedy these limitations, we propose the creation of a task simulator and the use of Symbolic Regression to learn expressions relating model performance to possible factors of influence. For MTL, we study the model performance against the number of tasks (T), the number of samples per task (n) and the task relatedness measured by the adjusted mutual information (AMI). In our experiments, we could empirically find formulas relating model performance with factors of $\sqrt{n}$, $\sqrt{T}$,  which are equivalent to sound mathematical proofs in \cite{Maurer2016}, and we went beyond by discovering that performance relates to a factor of $\sqrt{AMI}$. 
\end{abstract}

\section{Introduction}

Deep learning methods have achieved state-of-the-art performance in many applications when learning to solve a single problem. In this domain, current efforts are partly focused on engineering new neural architectures or exploring novel methods to improve the accuracy of these single-task models. There is another strain of research that explores learning models that share a certain degree of knowledge among different related tasks. A common approach is to use multitask learning (MTL).

In this setting, two or more related tasks are jointly trained such that it is expected that the model achieves better generalization for all the tasks. The main assumption is that related tasks might contain complementary information, which should act as an inductive bias to guide the model towards a better optimum as compared to training one model for each task in isolation \cite{Caruana1997}.

Multitask learning has been applied to many problems over time, particularly in NLP. \citet{Collobert2008} proposed a unified architecture to mutually learn different sequence labeling tasks, a language modeling task and a semantically related words problem. \citet{Mccann2018} used an MTL-based model to solve 10 different tasks in natural language (including semantic parsing, machine translation, summarization and sentiment analysis) by casting them as question answering. \citet{Crichton2017} proposed an MTL-based Named Entity Recognition (NER) in biomedical text mining. In Biomedicine, there are a variety of named entities datasets for the different sub-domains (e.g. genes, proteins, chemicals and species names). Some of these datasets are very small, which harms the performance of a model trained to classify named entities. However, obtaining labeled datasets is usually done manually by experts and are expensive to develop. Therefore, \citet{Crichton2017} cast domain-specific NER datasets as the different tasks and proposed a model to mutually learn them to improve generalization. Their results disclose yet another benefit of MTL models, by learning multiple tasks the model capitalizes from more data, such that tasks with small training sets can generalize better.

Even though great average improvements were achieved, the results are often mixed when evaluating each task in isolation and comparing them with the state-of-the-art performance of single-task models \cite{Collobert2008, Crichton2017, Sogaard2016, Bingel2017, Changpinyo2018}. The challenge of selecting a set of tasks, which guarantees better generalization for all tasks in multitask settings, remains partially unsolved. This is, to a certain extent, due to the few theoretical investigations that clearly expose the conditions under which multitask learning leads to better performance.

The few works that do try to theoretically understand generalization in MTL, however, rely mostly on strong assumptions about how tasks are related or are only applied to simple scenarios, such as settings where all tasks contribute with datasets of the same size. \citet{Baxter2000} proposed generalization bounds for MTL using VC-dimension and covering numbers, however, he assumes that task relatedness is given a priori. \citet{Maurer2006} used the Rademacher complexity to analyze linear multitask methods. In a later work, \citet{Maurer2016} have expanded their evaluation to nonlinear methods and assumed that similar tasks shared a common feature representation. They proposed a bound based on the Gaussian average while evaluating their theoretical results for noiseless binary classification tasks.

The mixed results obtained in MTL motivate to study models that are more theoretically grounded while avoiding being trapped into prior assumptions and constrained problems. In this work, we skip strictly deriving bounds for a well-defined domain and rather follow the scientific methodology that inspired Johannes Kepler to be the first to correctly explain the elliptical orbit of planets amongst other riddles involving planetary motion. In his work, Kepler applied the scientific methodology of using observational data to discover functions relating variables of interest. The laws empirically obtained by Kepler were later mathematically proven by the theoretical works of Isaac Newton.

Following an approach similar to Kepler's, we will try to learn formulas from data by using a method called Symbolic Regression. The empirically obtained expressions are able to explain the generalization performance of multitask models with respect to given parameter(s) (e.g. the number of tasks and task relatedness). With that, we also expect to find a more elementary way to group tasks in multitask models and elucidate the benefits of MTL.

\paragraph{Our Contributions}
\begin{itemize}
\item We developed a task simulator to create a multitask environment of sequence labeling tasks. This simulator enables the user to control the number of tasks, the level of relatedness between the tasks, and the number of samples per task.
\item We used a novel approach based on Symbolic Regression to study the convergence patterns of machine learning algorithms according to parameters of interest. More specifically, we apply this methodology to the non-trivial case of multitask learning models for sequence labeling tasks. 
\item  We discovered simple formulas that quantify the performance of MTL models as a function of the controlled parameters enabled by the task simulator. 
\end{itemize}

\section{Background and Related Work}

In the following, we supply the background information and related literature to formally introduce the main concepts essential to this work.

\paragraph{Multitask Learning} 
Multitask learning is a method that aims to achieve better generalization by learning $T$ related tasks jointly. The improvement in MTL happens because if tasks are related, features that are learned for one task might benefit the learning of some other related task. Multitask learning also has an implicit data augmentation effect, assuming that different tasks might contribute with their own data set. MTL will only induce better generalization if the objectives of each task are not conflicting, which is usually the case when tasks are related. Therefore, MTL is not robust when facing outlier tasks, where performance can degrade quickly \cite{Lee2017}.

\citet{Caruana1997} describes multitask learning as a form of inductive bias. He argues that learning one task at a time can be counterproductive because it misuses the additional knowledge that training signals from different tasks in the same domain usually have. Training signals from one task can improve a model by introducing a bias when learning some other task which will cause the whole model to prefer better hypotheses. The inductive bias behaves similarly to regularizers. It prevents the model from overfitting since it is not trying to optimize a single goal, but multiple objective functions for the different tasks. 

\citet{Maurer2016} used the Empirical Risk Minimization (ERM) framework to derive their theoretical results. A general formulation of ERM for multitask learning settings adds the notion of optimizing for multiple tasks jointly. Given $T$ tasks $\{t_{1}, t_{2}, ..., t_{T}\}$, where each task has its own dataset with samples of the form $z_t = (x_{t,i}, y_{t,i})$. Suppose we are given $n$ independent and identically distributed samples for each task and these samples are generated from unknown distributions $p_{t}(z)$. Let $Q(z_{t},w_{t})$ be a set of loss functions of the form $L(y_{t},f(x_{t},w_{t}))$ and $w_{t} \in \Omega$. Then, the goal is:

\begin{equation}
\min_{w_{1}, ... w_{T}}  \frac{1}{T} \sum_{t=1}^{T} \frac{1}{n} \sum_{i=1}^{n} L(y_{t,i},f(x_{t,i},w_{t}))
\end{equation}    

\paragraph{Bounds and Convergence Rates for Multitask Learning}
\citet{Baxter2000} derived theoretical bounds for MTL based on the VC-dimension and covering numbers theory under the assumption that tasks from the same environment are related. The set of tasks from the same environment was assumed to be probabilistically distributed. He was able to find bounds of the order $O(\sqrt{\log{n}/n})$ and $O(\sqrt{\log{T}/T})$, where $n$ is the number of samples per task and $T$ is the number of related tasks.

Other convergence rates for MTL were reported in the literature. The bound in \cite{Maurer2016} is based on Gaussian average and yields convergence rates of the order of $O(\sqrt{1/n})$ and $O(\sqrt{1/T})$. However, in this work, task relatedness is not explicitly explored. \citet{Liu2017}  explore the notion of algorithmic stability rather than using complexity measures to derive algorithm-dependent upper bounds. Their bounds achieve fast convergence rates of order $O(1/n)$  and of order $O(1/T)$. Another usual assumption of many theoretical studies in MTL is to encode a priori that the relations between the related tasks are linear. To overcome this limitation, \citet{Ciliberto2017} study the generalization properties of MTL with nonlinear task relations. They found that nonlinear MTL for a set of unrelated tasks converges at a rate of $\tilde{O}((T/n)^{1/4})$ and that for the set of related tasks, this rate is $\tilde{O}((1/nT)^{1/4})$.

\paragraph{Quantifying Task Relatedness in MTL}
The assumptions made about the level of relatedness between tasks is one of the key aspects to obtain successful results in MTL. Naively learning unrelated tasks together may be detrimental and outlier tasks may lead to negative transfers that can damage performance.  In non-neural MTL, some preliminary research assumed that task relatedness was given a priori. Some studies supposed that all tasks were uniformly related and were controlled globally by imposing that the Frobenius norm of the differences between parameters of all tasks is close to each other \cite{Evgeniou2004, Parameswaran2010}. Other studies adopted a pairwise approach, evaluating the similarity between all pairs of tasks. A pair of tasks are similar when the difference between the parameters of them is upper bounded by some constant, or when they would use features in a similar way \cite{Evgeniou2005}. The results from \cite{Evgeniou2005} were further explored, such that one could model clusters \cite{Kang2011} or hierarchies \cite{Goernitz2011} among tasks.  Methods for inferring task relatedness from data were also proposed. \citet{Bonilla2007} explored learning task relatedness using Bayesian models and \citet{Zhang2014} used a regularization approach. 

In the context of neural MTL for NLP, task relatedness is not as explored as for non-neural methods. Some works assume relatedness is given a priori so that relationships between tasks are linguistically motivated \cite{Sogaard2016, Sanh2019}. Recent works also adopt a brute-force approach by training all pairs of tasks to realize which pairs lead to gain and losses when jointly trained \cite{Bingel2017, Changpinyo2018}. Other methods are focused on neural architecture design. \citet{Changpinyo2018} include a task embedding to learn task similarity and \citet{Ruder2019} propose multitask architecture learning that lets the network itself learn which layers to share, the amount of sharing per layer and how to best design the output layer so as to maximize the benefit of mutually learning several tasks. Finally, \citet{Clark2019} rather focused on the multitask training algorithm. They use concepts of knowledge distillation such that a “teacher” model (single-task model) transfers knowledge to a “student” model (multitask model) by training the student to mimic the teacher’s outputs.

\paragraph{Symbolic Regression}
Symbolic Regression \cite{Orzechowski2018} is a learning method to find a formula that relates some input-output pair. A common implementation of Symbolic Regression is done using Genetic Programming (GP) \cite{Koza1992}. The building blocks for GP-based methods are constants, variables and algebraic operations. The terminal set is composed by the input variables and constants. The primitive function set consists of the relevant operators (e.g. op = \{$+, -,$ $\div$, $\times$, $inv, abs, cos, sin, log$\}). In general, an expression $ (log(x_{2})+0.5)x_{1}$ is represented as syntax tree or an S-expression, such as $ y = \times (+ (log (x_{2}), 0.5),x_{1}) $. The GP-based method for Symbolic Regression consists of 5 phases: creation of an initial population of random formulas, evaluation of the current population, selection of individuals, crossover, and mutation. For more information about each of these phases, one can refer to \cite{Koza1992}.

\paragraph{Sequence Labeling}
Many NLP tasks can be cast as a sequence labeling problem, where the goal is to assign a label to each token in a sequence. Sequence labeling tasks serve also as an intermediate step to achieve more complex problems, which are usually of more practical utility to understand language. As an example, question answering systems commonly have Named Entity Recognition (NER) as an embedded component. The goal of NER is to identify a word or a set of words as named entities (e.g person, organization, location, time and quantity).

Traditionally, sequence labeling problems were solved by probabilistic graphical models including Hidden Markov Models \cite{Rabiner1989}, Conditional Random Fields \cite{Lafferty2001} and Maximum Entropy Markov Models \cite{Mccallum2000}. These models heavily relied on hand-designed features carefully engineered for a certain task, language and dataset. \cite{Collobert2011} tackled these limitations by proposing neural-based methods. Instead of task-specific engineering, they let the network discover internal representations that are useful for learning a given task. More recently, Transformer based models \cite{Vaswani2017, Devlin2019} have been used as one of the state-of-the-art architecture for problems involving sequential data. 

\section{General Approach}

\begin{figure}[hbt!]
\centering
\includegraphics[width=0.95\linewidth]{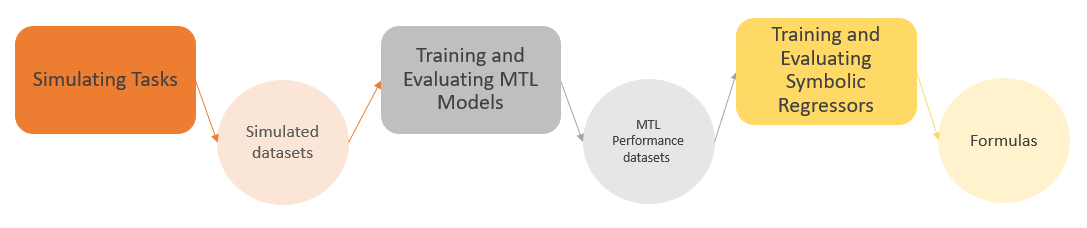}
\caption{Empirical methodology}
\label{fig:methodology}
\end{figure}

To tackle the challenge of quantifying the benefit of MTL, we propose the following methodology (visualized in Figure~\ref{fig:methodology}). First, we developed a task simulator to create data for $T$ artificial tasks while controlling parameters of interest, such as the number of tasks, the number of samples per task and task relatedness. Then, we trained and evaluated more than 10000 multitask models with varying architectures using the generated data from the task simulator. We record the performance of these models and the values of the controlled parameters into a new dataset. Using this new dataset, we apply Symbolic Regression to experimentally find simple formulas relating the empirical model performance and the controlled variables.

\paragraph{Task Simulator}
We developed a simulator to generate tasks under controlled conditions. This simulator allows a user to create a multitask environment while controlling:
\begin{itemize}
\item the task relatedness measured by the adjusted mutual information (proxied by a parameter $\alpha$)
\item the number of artificial tasks
\item the number of samples per task
\end{itemize}

Our task simulator uses as a starting point a real dataset (in our experiments from the NER domain). Based on this dataset, related tasks are generated. The user controls relatedness between pairs of the form \{RealTask, ArtificialTask$_{t}$\} using a matrix $M$ that regulates the relationship between the real task tag-set (with labels \{$r_{1}, r_{2}, .. , r_{k}$\}, where $k$ is the number of different labels in the real task), and a generated tag-set of synthetic labels (\{$s_{1}, s_{2}, .. , s_{c}$\}, where $c$ is the number of different labels in the synthetic task). Each entry of $M$ represents conditional probabilities of the form $P(ArtificialLabel = s_{i} \mid RealLabel = r_{j})$, and is defined as:
\begin{align*}
M [i, j] = 
\begin{cases}
    \alpha & \text{if } i = j\\
    \frac{1 - \alpha}{c - 1}   & \text{otherwise}
\end{cases}
\end{align*}

where $1 \leq i \leq k$ and  and $1 \leq j \leq c$. $\alpha$ is a parameter that controls how much probability mass is kept in the diagonal of the matrix. Hence when $\alpha = 1$ we have the maximum dependence between the tasks and when $\alpha = 1/c$, we can say that the tasks are independent. The parameter $\alpha$ is a proxy for the Adjusted Mutual Information (AMI) as defined in Equation ~\ref{eq:ami}.

\begin{equation}
AMI(U,V) = \frac{I(U,V) -  E\{I(U,V)\}}{max\{H(U),H(V)\} - E\{I(U,V)\}} \label{eq:ami}
\end{equation}

In the equation above, $H(U)$ and $H(V)$ are the entropy of two clusterings U and V, respectively. E\{I(U,V)\} is a factor to correct the original Mutual Information (MI) metric for randomness. The Adjusted Mutual Information quantifies how much information is shared by two random clusters. This quantity is symmetric $AMI(U,V) = AMI(V,U)$ and non-negative, where  $AMI(U,V) = 0 $ means that $U$ and $V$ are independent. This metric is always in the interval [0,1]. We omit further details about this adapted measure and rather refer to the work \cite{Vihn2010} for additional information.

We build several of these matrices in order to simulate artificial tasks with different levels of task relatedness. For simplicity, we create artificial tasks with the same number of labels in the tag-set of the real task. Algorithm 1 further outlines the implementation details about the task simulator.

\begin{algorithm}[H]
 \SetAlgoLined
 \KwData{Array of Real Task $\var{R}$, Number of Tasks $\var{T}$, Number of Samples per Task $\var{n}$,Array of $\var{alpha}$}
 \KwResult{Matrix of $T$ Synthetic Tasks $\var{SYN}$, Array of Adjusted Mutual Information \var{AMI}}
 $\var{LabelSet} \leftarrow GetLabelSet(\var{R})$ \;
 \For{t = 1 ...T iterations}{
 $\var{M} \leftarrow GenerateLabelMatrix(\var{LabelSet}, \var{alpha[t]})$ \;
 \For{i = 1 ... n iterations}{
 $\var{probabilities} \leftarrow \var{M[R[i]]}$ \;
 $\var{SYN[i][t]} \leftarrow  RandomChoice(\var{LabelSet}, \var{probabilities})$ \;
 $\var{AMI[t]} \leftarrow  AdjustedMutualInformation(\var{R}, \var{SYN[t]})$ \;
}
}
 \caption{Task Simulator}
\end{algorithm}

\begin{figure}[hbt!]
\centering
\includegraphics[scale = 0.75]{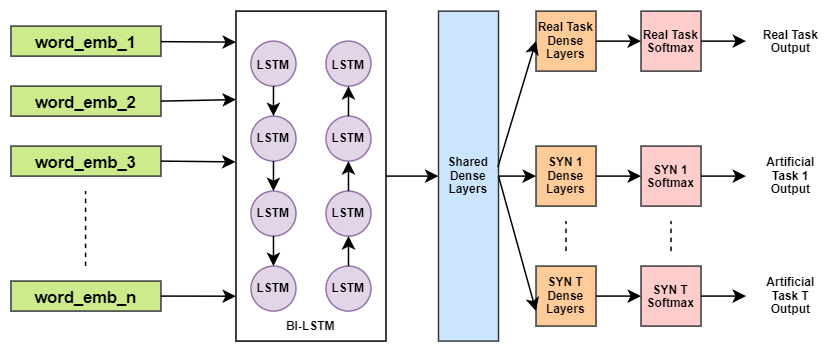}
\caption{Model architecture}
\label{fig:architecture}
\end{figure}

\paragraph{Architecture of MTL Models}
In an attempt to discover formulas that are not particular to one specific model, we created different MTL architectures as a combination of a shared BI-LSTM layer and fully connected layers  (see Figure ~\ref{fig:architecture}). The difference in each architecture lies in the number of dense layers shared or task-specific in each model. Initially, the input is fed to a shared BI-LSTM layer, which will then learn generic parameters that are available for all tasks during training. Then, the output of BI-LSTM flows into fully connected layers that can be either shared or task-specific. In a multitask setting, the output layers are always task-specific.

\paragraph{Applying Symbolic Regression} By training and evaluating the MTL models on the multitask datasets generated by the task simulator, we obtain a model performance for each dataset (e.g. $F_{1}score$). These can be used to obtain new, different datasets on which the symbolic regressors can be fit. These datasets follow the format [$I_{1}, I_{2}, ..., I_{p}, Out$], where \{$I_{1}, I_{2}, ..., I_{p}$\} are the controlled factors (e.g. \{$n, T ,AMI$\}) and $Out$ is the measure of model performance (e.g. $F_{1}score$). Using datasets of this format, we were able fit formulas with different number of parameters, including $Out= f(I_{p})$ and $Out= f(I_{1}, I_{2}, ..., I_{p})$. We evaluated the obtained formulas with the Mean Squared Error (MSE) by using held-out datasets.

\section{Experimental Setup}

For the purpose of this work, we adapted the simulator to output artificial sequence labeling tasks that are related to the Named Entity Recognition problem.

\paragraph{Training and Evaluating MTL models} All models were implemented and trained using Pytorch \cite{Pytorch}. The input for all models was the words of the CoNLL dataset \cite{CoNLL} embedded with the pre-trained FastText vectors available in \cite{fasttext}. The input words were converted to sentences of the form [Left Context, Target Word, Right Context]. The size of the context window was set to 3. The output has the format [NER label, Synthetic Task$_{1}$ label, ... , Synthetic Task$_{t}$ label] corresponding to the labels for the center word on $t+1$ tasks. We converted the encoding of the original CoNLL dataset from the IOB tagging scheme to IO encodings for the training set and IOB2 for the development and test sets.

We adopted a similar version of the fully joint strategy in \cite{Mccann2018} to train our models. For comparability, we trained all models using the Adam Optimizer with default parameters over 100 epochs. The batch size is also fixed to 100. At each step, we backpropagate the total cross-entropy loss defined as the sum over the losses for all tasks in the related set. To evaluate the accuracy of our models we used the official CoNLL evaluation script \cite{evalscript}. In this work, we report only the overall $F_{1}score$ metric for the Named Entity Recognition task. Each $F_{1}score$ reported is an average over 5 runs.

\paragraph{Fitting Formulas} To find the formulas we used the gplearn package \cite{Stephens2019}. This package performs Symbolic Regression using a GP-based implementation.

The formulas were obtained from datasets created for fitting functions of the form $F_{1}score = f(T, n, AMI)$. We used a 90/10 train/test split for each experiment. We chose a set from $\{+ , - , \div , \times , \log , \sqrt , exp , neg\}$ to represent the primitive set of operators. The exponential operator was not available in the library, therefore, we had to add it manually. We train all models for at least 20 generations while controlling the diversity of formulas at each generation under a 70\% probability of crossover and 10\% probability of point, hoist and subtree mutation. We set the parsimony coefficient to 0.2 in most experiments. This last hyperparameter controls the tradeoff between fitness and expression length. Finally, we fit the formulas using the actual values of X, but a scaled y according to $c*y$, where c is a constant which was set to 100 and 1000 in the different experiments. We have done this because, in our experience with the gplearn module, when y values vary from [0,1], the formulas always converged to a constant close to the mean value of y.

\section{Experimental Results}

It is our intention to study the benefit of multitask learning by adopting an empirical approach. Albeit the references mathematically exploring the advantage of MTL, we take an empirical approach other than theoretical methodologies and sound numerical proofs. We do that for two reasons. First, we wanted to verify if a purely empirical tool, such as Symbolic Regression, would find similar results already theoretically explored. Secondly, we wanted to avoid being trapped by prior assumptions but rather letting a learning algorithm discover explanatory patterns by itself.

We conducted experiments to inspect the patterns of model performance when considering the influence of (1) sample size per task; (2) adjusted mutual information; (3) number of tasks; and (4) all these parameters together. The results of our experiments were used to induce simple expressions to quantify the advantage of MTL models trained under these variable conditions. Two scenarios were explored. First, we performed experiments with balanced datasets, where all tasks have the same number of samples, which is a common assumption in theoretical studies. Secondly, we explored the more realistic scenario where tasks have datasets with a different number of samples.

\subsection{Experiments with Balanced Datasets}
To fit the functions we create the MTL Balanced Datasets which contains 1002 observations on 5 features (n, T, AMI, $\alpha$, $F_{1}score$ ). Each row in our dataset represents the average $F_{1}score$ of 5 models trained with the same variables configuration. Hence, we trained and evaluated 5010 models to obtain this dataset. We vary $T \in \{1, 2, 3, 4, ... , 15\}$, $n \in \{1000, 1500, 2000, 2500, 3000, ...., 20000\}$, $AMI \in \{0, 0.13, 0.17, 0.21, 0.27, 0.34, 0.43, 0.55, 0.57, 0.6, 0.63, 0.68, 0.71, 0.75, 0.8, 0.85, 0.91, 1\}$ and $\alpha \in \{0.2, 0.6, 0.65, 0.75, 0.8, 0.85, 0.87, 0.9, 0.91, 0.92, 0.93, 0.94, 0.95, 0.96, 0.97, 0.98, 0.99, 1\}$.

\begin{figure}[hbt!]
     \centering
     \begin{subfigure}[b]{.3\linewidth}
         \centering
         \includegraphics[width=\linewidth]{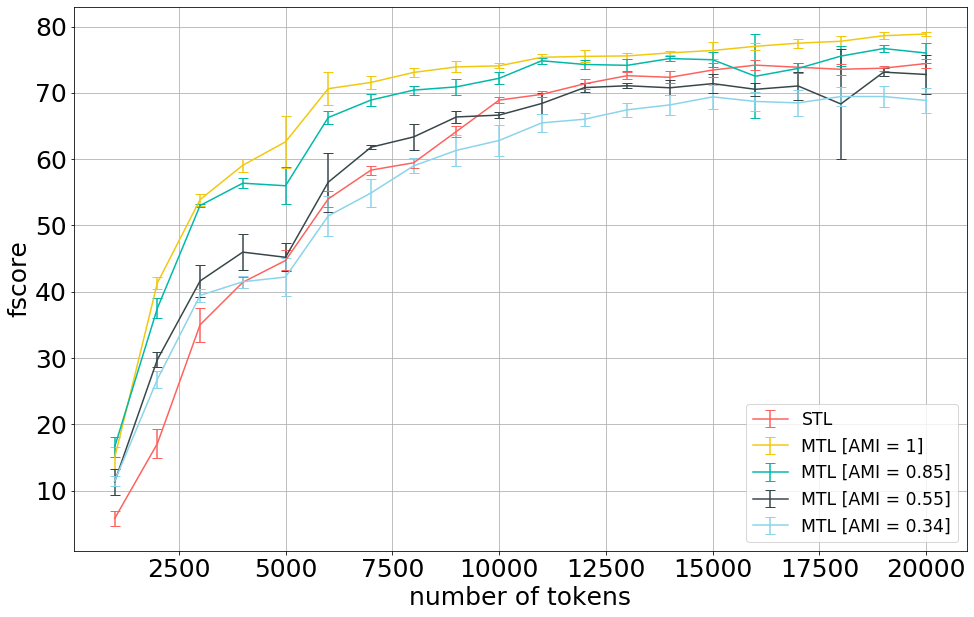}
         \caption{}
         \label{fig:data}
     \end{subfigure}
     \hfill
     \begin{subfigure}[b]{.3\linewidth}
         \centering
         \includegraphics[width=\linewidth]{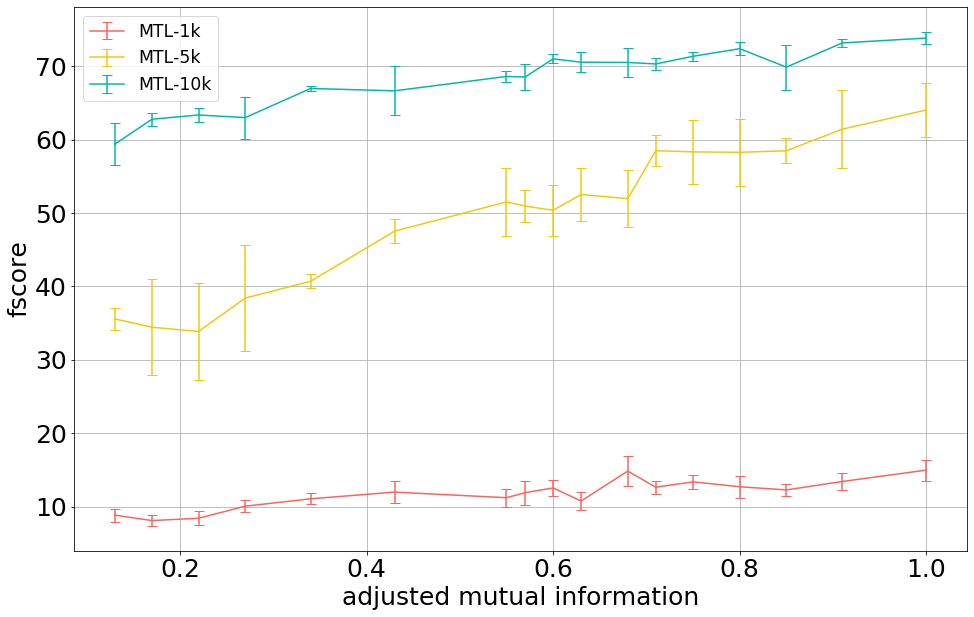}
         \caption{}
         \label{fig:mi}
     \end{subfigure}
     \hfill
     \begin{subfigure}[b]{.3\linewidth}
         \centering
         \includegraphics[width=\linewidth]{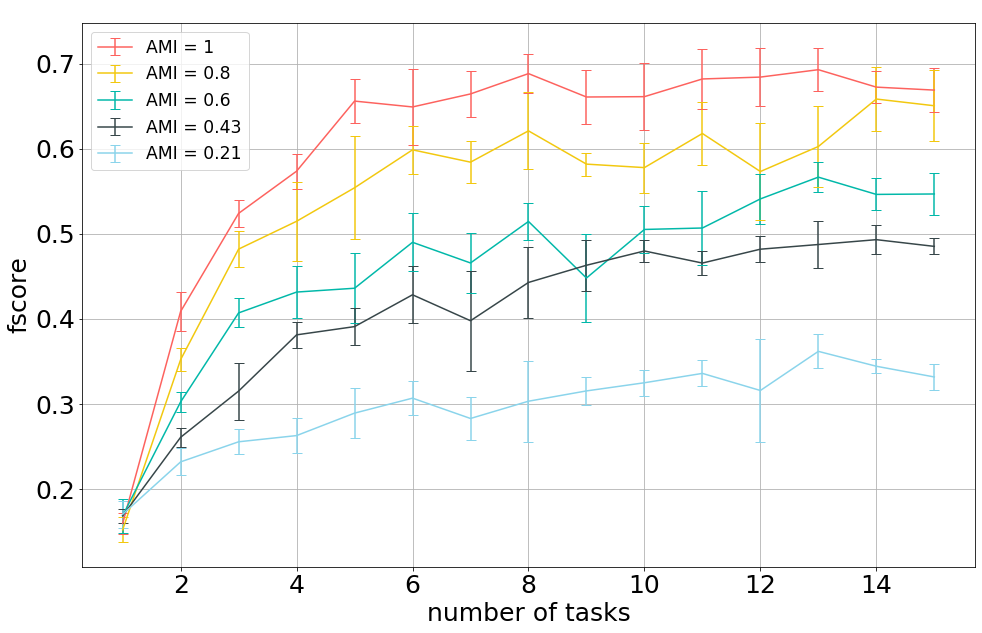}
         \caption{}
         \label{fig:task}
     \end{subfigure}
        \caption{Results for experiments with balanced datasets}
        \label{fig:exp1}
\end{figure}

\paragraph{Considering Number of Samples per Task}
We study the effect of the number of samples per task on the $F_{1}$score of the NER task for single task (STL) models and MTL models. We vary the number of tokens per task from 1k up to 20k samples for both settings. Figure ~\ref{fig:data} shows the performance of the STL and 4 MTL models ($T=2$), with different amounts of adjusted mutual information. For comparability, we only plot MTL models with the same number of tasks ($T=2$). The plot highlights the positive impact of the increased sample size on both STL and MTL settings, where MTL outperforms STL models when AMI between the related tasks is high. 

We also observe that all models have a similar curve shape when varying this parameter, which characterizes a good scenario for fitting symbolic regressors. We fit a symbolic regressor to discover a function of the form $F_{1}score = f(n)$, where $n$ is the number of tokens per task. The regressor returns the following function with a test MSE of 0.027.
\begin{equation}
F_{1}score = \sqrt{n}*0.69 \label{a}
\end{equation}

\paragraph{Considering Adjusted Mutual Information}
In this experiment, we investigate the impact of the adjusted mutual information on the performance of the NER task. We vary the AMI from 0.13 to 1. Each color in Figure ~\ref{fig:mi} represents a MTL model with different amounts of samples per task. From this plot, we note that for all multitask scenarios, we obtained improvement in generalization as the AMI in the task set increases. In this experiment, we obtained empirical evidence that the benefit of multitask learning is almost certain to be observed in low resource scenarios even for low task relatedness. This advantage is still observed for high resource settings, but the task set needs a higher AMI value.

Once more we found a function of the form $F_{1}score = f(AMI)$. The formula below has a test MSE of 0.034. 
\begin{equation}
F_{1}score =  \sqrt{AMI} / 0.015 \label{b}
\end{equation}   

\paragraph{Considering Number of Tasks}
We vary the number of tasks up to 15 and evaluate the impact in the performance of the NER task. Each task compromised 2k tokens. Figure ~\ref{fig:task} contains different MTL models with varying levels of adjusted mutual information. Evaluating this plot, we can conclude that all MTL models outperform their STL equivalent and that not always the multitask models benefit from adding more tasks.

In this experiment, symbolic regression failed to converge to a meaningful formula. By observing the shape of the curve and the results from the previous experiment, we guessed factors of  $\sqrt{T}$ as a fitting formula. We manually tried a few constants and report the expression with the lowest test MSE of 0.072.
\begin{equation}
F_{1}score = \sqrt{T}*20 \label{c}
\end{equation}

\paragraph{Considering All Parameters}
In this experiment, we wanted to obtain an expression with all three factors previously studied. Instead of searching for formulas of the form $F_{1}score = f(n)$, where the input was a unique variable, our goal was searching for an expression $F_{1}score = f(n, AMI, T)$. We report the simplest formula with the best fit among all outcomes from the symbolic regressor. The best candidate expression with the lowest test MSE of 0.009 is reported below. 
\begin{equation}
F_{1}score =  \sqrt{(log(T) + 0.45) * (AMI * n/(AMI +0.6)))} \label{d}
\end{equation}   

\paragraph{Discussion} When we evaluate the 4 formulas obtained in this set of experiments, we could infer convergence rates of:
\begin{itemize}
\item $O(\sqrt{n})$ in the number of samples based on the Equations ~\ref{a} and ~\ref{d},
\item $O(\sqrt{AMI})$ in the amount of adjusted mutual information from the Equations ~\ref{b} and ~\ref{d},
\item $O(\sqrt{T})$ from Equation ~\ref{c} and $\sqrt{(log(T)}$ from Equation ~\ref{d} in the number of tasks.
\end{itemize}

While contrasting with previous theoretical work, we conclude that our results are not far off the outcomes from sound mathematical proofs adopted in \cite{Maurer2016, Baxter2000, Ciliberto2017, Liu2017}. In fact, the convergence rates from Equations ~\ref{a} and ~\ref{c} are equivalent to \citet{Maurer2016}. Besides, we were able to derive novel results relating model performance with the adjusted mutual information according to Equations ~\ref{b} and ~\ref{d}.

\subsection{Experiments with Unbalanced Datasets}
In this second set of experiments, we explore the scenario where tasks contribute with datasets of different sizes. We generated data where the NER task has always less data than the synthetic tasks.  We created the MTL Unbalanced Datasets which consists of 1555 observations on 6 features ($n_{ner}$, $n_{syn}$, $T$, $AMI$, $\alpha$, $F_{1}score$). We vary $T \in \{1, 2, 3, ..., 30\}$, $n_{ner} \in \{1000, 1500, 2000, 2500\}$, $n_{syn} \in \{2000, 3000, ..., 6000\}$,  $AMI \in \{0.13, 0.22, 0.43, 0.55, 0.61, 0.71, 0.8, 0.91\}$ and $\alpha \in \{0.6, 0.7, 0.85, 0.9, 0.92, 0.95, 0.97, 0.99\}$ . We performed the same set of experiments as for the balanced datasets and the resulting formulas were: 

\begin{equation}
F_{1}score =  \sqrt{n_{ner}}
\end{equation}   
\begin{equation}
F_{1}score =  \sqrt{n_{syn}}*0.69
\end{equation}   
\begin{equation}
F_{1}score =  \sqrt{2*AMI} / 0.026
\end{equation}   
\begin{equation}
F_{1}score =  log(\sqrt{T}+ log(T))/ 0.039
\end{equation}    
\begin{equation}
F_{1}score =  (\sqrt{log(n_{ner}/0.38)*(\sqrt{T}+T)})*\sqrt{n_{syn}*AMI}
\end{equation}  

\section{Conclusion}

In this work, we developed a simulated environment of sequence labeling tasks to explore the complexities of learning multiple tasks jointly. Using the data generated from a task simulator, we were able to investigate the generalization patterns of MTL models. Further, we resorted to a learning algorithm to obtain symbolic expressions that can assist practitioners to understand the factors which contribute to the generalization performance of MTL models in the sequence labeling domain. Our findings were in accordance with previous theoretical works, yet we were able to add additional insights to the unexplored setting of unbalanced datasets. In addition to that, we explicitly quantified task relatedness rather than making any assumption about it or trying to learn it implicitly as a parameter of the model. 

\section*{Broader Impact}

We bring awareness about the use of multitask learning and how to understand its benefits using a novel methodology. Approaching theoretical questions via simulation and the learning of formulas from data could ease the way towards more explainable machine learning processes.

The possible negative impact of this work is the idea that our methodology could be seen as a replacement instead of a support of theoretically grounded work. Undoubtedly, rigorous proofs are an essential part of our domain. Theoretical contributions are necessary since they make the use of machine learning more reliable. Essentially, the only way one can safely say some phenomena is true is by presenting a valid mathematical proof for it.

Building a better understanding of machine learning algorithms, and especially deep learning methods, is an important requirement for applying them in a responsible manner in real applications. More specifically, we believe that knowledge sharing techniques such as MTL and Transfer Learning are important components of many low-resource problems. Improving the use of these techniques enables the use of machine learning for unsolvable problems due to a lack of data. These problems include, for instance, the use of natural language processing for languages in developing countries or general classification tasks for small businesses. 

\section*{Acknowledgments}

Gefördert durch die Deutsche Forschungsgemeinschaft (DFG) – Projektnummer 232722074 – SFB 1102 / Funded by the Deutsche Forschungsgemeinschaft (DFG, German Research Foundation) – Project-ID 232722074 – SFB 1102 and the EU-funded Horizon 2020 project ROXANNE under grant number 833635.

\bibliography{literature}

\end{document}